\documentclass{article}

\PassOptionsToPackage{numbers, compress}{natbib}

\usepackage[preprint]{neurips_2023}

\usepackage[utf8]{inputenc} 
\usepackage[T1]{fontenc}    
\usepackage{hyperref}       
\usepackage{url}            
\usepackage{booktabs}       
\usepackage{amsfonts}       
\usepackage{nicefrac}       
\usepackage{microtype}      
\usepackage[dvipsnames,table,xcdraw]{xcolor}
\usepackage[font=small]{caption}

\hypersetup{
    colorlinks=true,
    linkcolor=BrickRed,
    citecolor=RoyalPurple,
    filecolor=magenta,      
    urlcolor=RoyalBlue,
}

\usepackage{graphicx}
\usepackage{bbm}
\usepackage{amsmath}
\usepackage{adjustbox}
\usepackage{textcomp}
\usepackage{makecell}
\usepackage{array,multirow,graphicx}
\usepackage{comment}

\usepackage[capitalize]{cleveref}
\crefname{section}{Sec.}{Secs.}
\Crefname{section}{Section}{Sections}
\Crefname{table}{Table}{Tables}
\crefname{table}{Tab.}{Tabs.}

\newlength\savewidth

\usepackage[misc]{ifsym}
\newcommand\blfootnote[1]{%
\begingroup
\renewcommand\thefootnote{}\footnote{#1}%
\addtocounter{footnote}{-1}%
\endgroup
}

\title{VLAB: Enhancing Video Language Pre-training \\by Feature Adapting and Blending}

\author{
    Xingjian He$^{1\,*}$ \ \
    \enskip Sihan Chen$^{1\,*}$\ \
    \enskip Fan Ma$^{2\,*}$ \ \
    \enskip Zhicheng Huang$^{3\,*}$ \ \
    \enskip Xiaojie Jin$^{4\,*\,}$\textsuperscript{\Letter} \\
    \enskip \textbf{Zikang Liu}$^{1}$
    \enskip \textbf{Dongmei Fu}$^{3}$ 
    \enskip \textbf{Yi Yang}$^{2}$ 
    \enskip \textbf{Jing Liu}$^{1\,}$\textsuperscript{\Letter}
    \enskip \textbf{Jiashi Feng}$^{4\,}$\textsuperscript{\Letter} \\
$^1$ The Laboratory of Cognition and Decision Intelligence for Complex Systems,\\ Institute of Automation, Chinese Academy of Sciences \\ \quad $^2$ Zhejiang University
 \quad $^3$ University of Science and Technology Beijing \quad $^4$ Bytedance Inc.
}

\begin{document}

\blfootnote{$^{*}$Equal contribution. Work done when Xingjian He, Sihan Chen, Fan Ma and Zhicheng Huang interned at Bytedance Inc.  \newline \hspace*{1.9em}{\scriptsize \Letter} Corresponding author: Xiaojie Jin$<$\url{jinxiaojie@bytedance.com}$>$, Jing Liu$<$\url{jliu@nlpr.ia.ac.cn}$>$, Jiashi Feng$<$\url{jshfeng@bytedance.com}$>$ \}}
\maketitle

\begin{abstract}
Large-scale image-text contrastive pre-training models, such as CLIP, have been demonstrated to effectively learn high-quality multimodal representations. However, there is limited research on learning video-text representations for general video multimodal tasks based on these powerful features. Towards this goal, we propose a novel video-text pre-training method dubbed VLAB: \textbf{V}ideo \textbf{L}anguage pre-training by feature \textbf{A}dapting and \textbf{B}lending, which transfers CLIP representations to video pre-training tasks and develops unified video multimodal models for a wide range of video-text tasks. Specifically, VLAB is founded on two key strategies: feature adapting and feature blending. In the former, we introduce a new video adapter module to address CLIP's deficiency in modeling temporal information and extend the model's capability to encompass both contrastive and generative tasks. In the latter, we propose an end-to-end training method that further enhances the model's performance by exploiting the complementarity of image and video features. We validate the effectiveness and versatility of VLAB through extensive experiments on highly competitive video multimodal tasks, including video text retrieval, video captioning, and video question answering. Remarkably, VLAB outperforms competing methods significantly and sets new records in video question answering on MSRVTT, MSVD, and TGIF datasets. It achieves an accuracy of 49.6, 61.0, and 79.0, respectively. Codes and models will be released.
\end{abstract}

\section{Introduction}
\vspace{-0.1cm}

In recent years, video language pre-training (VLP) has attracted growing interests by significantly advancing various video multimodal tasks \cite{li2022align,fu2021violet,wang2022git,wang2022all,bain2021frozen,li2022lavender,xue2022clip,zellers2021merlot,luo2020univl}, e.g. video text retrieval, video captioning, and video question answering (VQA). However, VLP faces substantial challenges like limited high-quality video-text data and huge training costs, which impede its further progress. Meanwhile, large-scale image language pre-training methods like CLIP~\cite{Radford2021LearningTV} have achieved remarkable success, showcasing their effectiveness in learning high-quality visual multimodal representations through extensive works~\cite{wang2022git,xue2022clip,gao2021clip2tv,tang2021clip4caption,zhong2022regionclip,luo2022clip4clip}. Inspired by this, a potential strategy involves transferring CLIP's learned representations to VLP for enhancing performance and training efficiency.

Nonetheless, transferring from image-text models to video-text models presents three primary challenges: domain, task, and data. 
 The first challenge is that these models target different domains. The second challenge is that while CLIP mainly deals with contrastive tasks, a foundational VLP model should also be equipped to handle generative tasks such as video captioning and video question answering. The third challenge lies in the pre-training data for these models, which varies significantly due to the availability of open-source data. Prior research has mainly focused on adapting CLIP for specific tasks such as video text retrieval~\cite{luo2022clip4clip,fang2021clip2video,xue2022clip} and video captioning~\cite{tang2021clip4caption,tewel2022zero}, but they do not address all tasks within a unified model. Recent studies have shown that directly training CLIP on video-text data fails to yield anticipated performance~\cite{xue2022clip,luo2022clip4clip}. This leads us to ask the question: \textit{how can we harness the representations of CLIP to create a unified and strong model for video language understanding?}

\begin{figure*}[t]
\begin{center}
\includegraphics[width=1\linewidth]{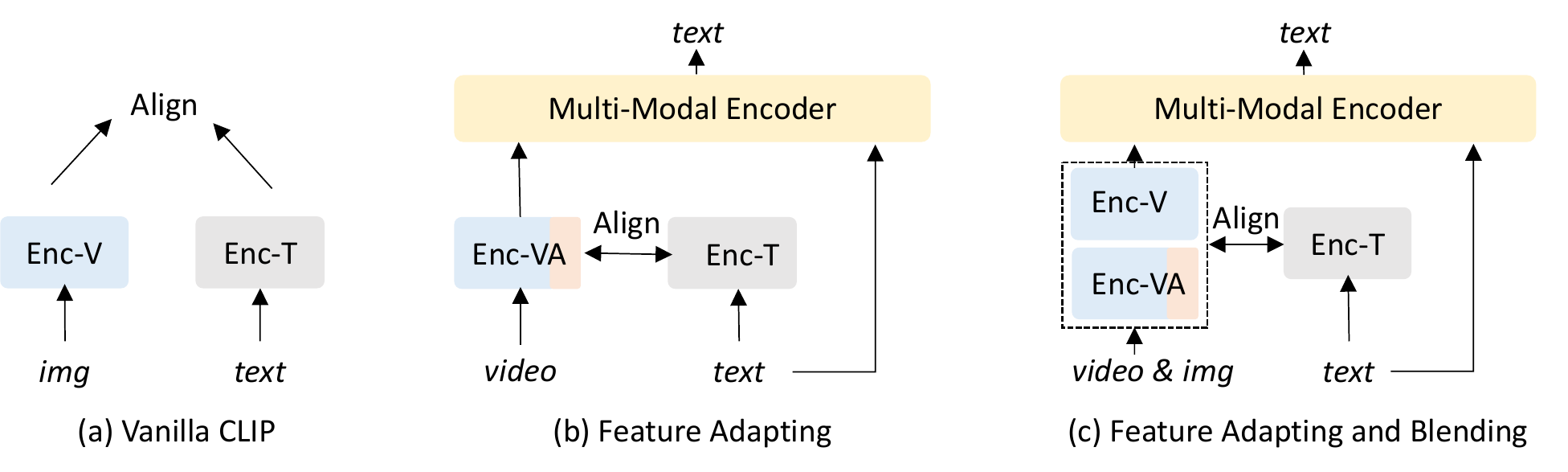}
\end{center}
\vspace{-0.2cm}
   \caption{VLAB Overview. Powered by the proposed feature adapting and feature blending approaches, VLAB presents \textbf{(c)} a unified video language model for handling both contrastive and generative tasks. ``Enc-V/T/VA'' denote the vision/text/adapted encoders respectively.}
\label{fig:overview}
\vspace{-13pt}
\end{figure*}

We propose VLAB, a novel video language pre-training method that transfers CLIP’s learned representations to video-text tasks. VLAB evolves from vanilla CLIP through feature adapting and blending. Feature adapting transforms CLIP features to the video domain via unifying generative and contrastive tasks, resulting in an intermediate model denoted as VLAB$_{\text{FA}}$ (depicted in \cref{fig:overview}(b)).  Temporal dynamics within videos can be easily modeled using it, thus overcoming CLIP's shortcomings in capturing temporal information. 
In order to address the problem of forgetting when adding new modules to the pre-trained CLIP~\cite{Radford2021LearningTV}, our approach involves adaptive transferring and integrated tuning. This fosters a more effective collaboration between the CLIP weights and adapter weights, ultimately leading to improved representations of videos. 

We further enhance the representations of temporally-aware model VLAB$_{\text{FA}}$ by integrating advantages from image and video domains, resulting in an improved model VLAB (see \cref{fig:overview}(c)). The final model uses a multimodal encoder to blend features learned in VLAB$_{\text{FA}}$ and CLIP’s image features. The feature blending strategy enables VLAB to automatically learn the optimal pattern for representation fusion. Experimental results demonstrate that this approach significantly improves the model’s potential for video-text understanding.

To validate VLAB's effectiveness, we conduct experiments on 8 widely-used benchmarks, spanning a variety of video multimodal tasks, including video QA, video captioning and text-to-video retrieval. VLAB achieves competitive performance across all tasks. Notably, it outperforms previously state-of-the-art methods that employ much larger models and pre-training data, such as  Flamingo~\cite{alayrac2022flamingo} and GIT2~\cite{wang2022git}. For example, on the evaluation of video QA using 
 MSRVTT/MSVD/TGIF, VLAB with 1.6B parameters significantly surpasses GIT2 (5.1B parameters) by 4\%/2\%/4\% respectively.
In summary, our contributions are threefold.
\begin{itemize}
    \item We propose a novel pre-training method for video language, called VLAB, to investigate how to transfer image-text models to video-text tasks. This leads to a unified model that has a strong ability for understanding video language.
    \item 
    To enhance and transfer CLIP representations, we introduce two pivotal tactics: feature adapting and feature blending. Both of these strategies have been proven to be highly effective in improving the final performance of the model.
    \item 

    Our model showcases remarkable outcomes when it comes to generative and contrastive video language tasks like VQA, video captioning, and text video retrieval. The model surpasses prior state-of-the-art methodologies that utilize bigger models and more extensive training datasets.
\end{itemize}

\section{Related Work}

\subsection{Video Language pre-training}

Exploration of video-language pre-training has been on the rise as a result of the triumphs of image-language pre-training, as evidenced by numerous studies~\cite{jia2021scaling,yao2021filip,yuan2021florence,yang2022unified,wang2021simvlm,chen2022pali,wang2022image,wang2022ofa,cho2021unifying,bao2022vlmo,dou2022empirical,huang2020pixel}. Pre-trained models have demonstrated impressive transferability in various video-language tasks, including video captioning, video text retrieval, and video question answering. In the past, early approaches~\cite{sun2019videobert,zhu2020actbert,li2020hero,miech2019howto100m} relied on pre-extracted offline features, such as S3D \cite{2020End}, for video representation. However, current trends showcase the significant benefits of training end-to-end models on vast web-crawled data~\cite{zellers2021merlot,bain2021frozen} like ALPRO~\cite{li2022align} and VIOLET~\cite{fu2021violet}. The CLIP model provided a step forward by successfully representing both visual and textual features, leading to numerous approaches~\cite{luo2022clip4clip,gao2021clip2tv,zhao2022centerclip,fang2021clip2video} achieving promising outcomes in the video-language domain. A notable example is GiT~\cite{wang2022git}, which utilized CLIP's vision encoder as a feature extractor and pre-trained it on a billion-scale cross-modality dataset with language modeling objectives. 
Training the GiT model necessitates a substantial amount of data. Nonetheless, by giving priority to the comprehension and knowledge obtained from pre-existing image-text models like CLIP, it is possible to promptly create a unified model that grasps video-language. This article illustrates how insights from pre-trained CLIP model were employed to design a powerful video language model, resulting in substantial enhancements in several video-language tasks like video captioning, video question answering, and video text retrieval.

\subsection{Adapters}
With the growing size of models, transfer learning has become increasingly important\cite{rebuffi2017learning}. The adapter has emerged as a promising approach for learning new tasks and not affecting the original model parameters~\cite{hu2021lora, sung2022vl, panst}. In particular,~\cite{rebuffi2017learning} proposed using adapters to enable the model to adapt to different images, while~\cite{rebuffi2018efficient} explored an efficient domain-adapter method using adapters.~\cite{chenadaptformer} proposed the AdapterFormer structure based on the ViT structure.~\cite{jie2022convolutional} and~\cite{panst} applied adapters to video tasks using spatial invariance and temporal relationships. In the multi-modal field, adapter applications are just beginning, with previous work mainly focused on specific tasks such as classification tasks~\cite{chenadaptformer,hetowards,jie2022convolutional}. VL-adapter~\cite{sung2022vl} only applies adapters to the text branch. In contrast to these previous works, our approach involves the application of adapters to a range of multi-modal tasks and specifically targets the enhancement of video language representations across different datasets.

\section{Method}
\label{sec:method}

The architecture of VLAB is shown in \cref{fig:frame_work}(a). 
The visual and text modalities are initially encoded separately, and a contrastive loss is applied to align their embedding spaces. VLAB incorporates two vision encoders that extract both temporal-aware and image representations. The text encoder follows the text branch in CLIP to encode textual information. Subsequently, a multi-modal encoder is employed as a modality fuser, utilizing visual embeddings to modulate the text generative task through cross-attention layers. 

The entire training process involves first training an adapted model VLAB$_{\text{FA}}$ and then training the integrated final model VLAB by synergizing the temporal and spatial representations learned in VLAB$_{\text{FA}}$ and CLIP. Next, we introduce the core components in detail.

\subsection{Feature Adapting}
\label{sec:feature_adapting}
\begin{figure*}
\begin{center}
\includegraphics[width=\linewidth]{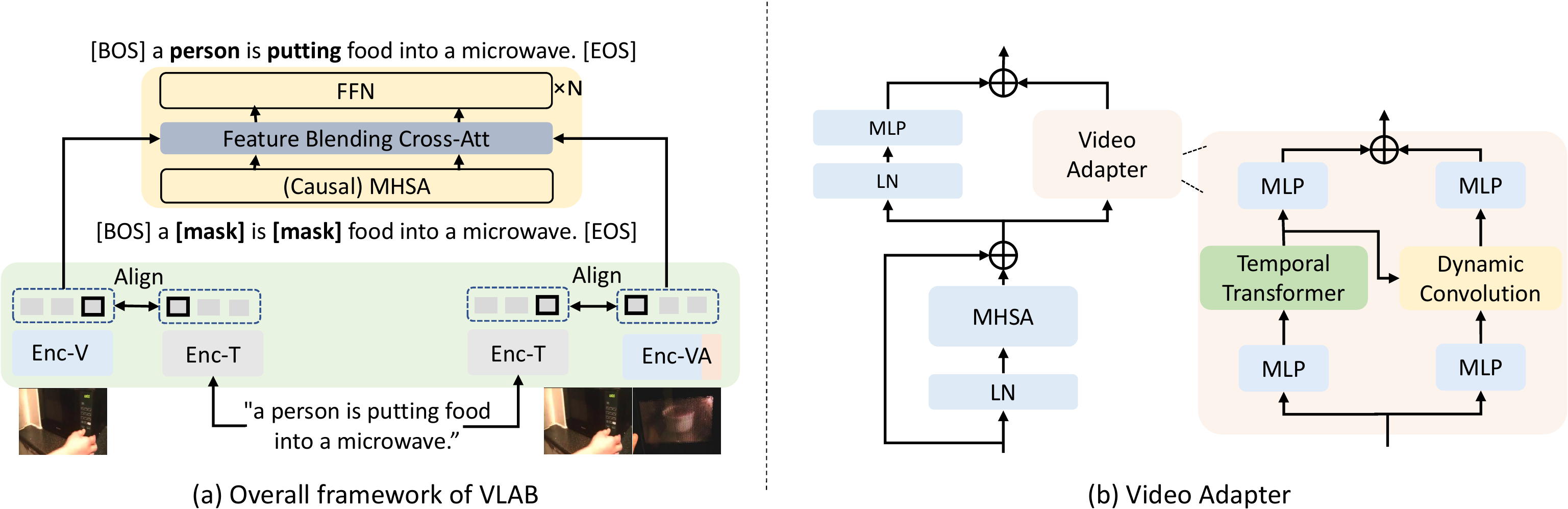}
\end{center}
   \caption{(a) Overall framework of VLAB. VLAB handles both contrastive and generative tasks in a unified model. (b) Illustration of the proposed video adapter module in feature adapting. Refer to Sec.~\ref{sec:method} for details.}
\label{fig:frame_work}
\end{figure*}

In this section, we introduce feature adapting, which aims to transform CLIP features into the video domain and unifies both generative and contrastive tasks, resulting in the VLAB$_\text{FA}$ model.
As depicted in \cref{fig:overview}(b), VLAB$_\text{FA}$ integrates a video adapter module within the CLIP visual encoder layer, allowing for the modeling of temporal dynamics in videos. Furthermore, it incorporates a multi-modal encoder on top of visual embedding, endowing it with text generation capability. 

\noindent\textbf{Video Adapter.}
As shown in \cref{fig:frame_work}(b), the video adapter has been placed inside the transformer block of the visual backbone, and its specifications are displayed on the right. 
The $l^{th}$ block contains the visual feature $\mathbf{v}_{l} \in \mathbb{R}^{N\times T\times d}$, where $N$ equals $N_p+1$. $N_p$ refers to the number of image patches accompanied by a global [CLS] token. $\mathbf{v}_{l}$ is denoted as $\{v^t_l\}_{i=1}^{|T|}$, where $T$ represents the number of frames in videos. The feature of the $t$-th frame, $v^t$, is in the form of $[v^t_{\rm [CLS]}, v^t_{\rm [patch]}]$. To represent the global frame features, we use $\mathbf{v}_{\rm [CLS]}=[v^0_{\rm [CLS]},\cdots,v^T_{\rm [CLS]}]$. Similarly, $\mathbf{v}_{\rm patch}=[v^0_{\rm patch},\cdots,v^T_{\rm patch}]$ represents the spatial patch features of the video.

To improve computational efficiency, we initially decrease the dimension of the global frame feature $\mathbf{v}_{\rm [CLS]}$ through an fully connected layer (\texttt{FC}). Next, we use a miniature transformer block, identical in structure to the one used in the visual backbone block, to consolidate the temporal information. As a result, the [CLS] token has temporal visual context encoding. Finally, we increase the feature's dimension by adding another \texttt{FC} layer, returning it to its original state.  The temporal aggregation can be written as:
\begin{equation}
    \tilde{\mathbf{v}}_{\rm [CLS]} = \texttt{FC}_2(\texttt{TT}(\texttt{FC}_1(\mathbf{v}_{\rm [CLS]}))),
\end{equation}
where $\texttt{TT}$ is the temporal transformer which is to absorb the temporal information from other frames. $\tilde{\mathbf{v}
}_{\rm [CLS]}$ is the updated [CLS] feature that encodes temporal visual context.

In addition, we incorporate spatio-temporal data into the patch features. To achieve this, we enhance the patch representation by utilizing the class token feature following the temporal transformer. Consequently, the patch features undergo an update process: 
\begin{equation}
\label{eq:patch_cal}
    \tilde{\mathbf{v}}_{\rm patch} = \texttt{FC}_4(\texttt{DyConv}(\tilde{\mathbf{v}}_{\rm [CLS]},  \texttt{FC}_3(\mathbf{v}_{\rm patch}))),
\end{equation}
where $\texttt{DyConv}$ is the dynamic convolutional operation as shown in \cref{fig:frame_work}(b) that applies a dynamic kernel from temporal [CLS] features on spatial patch features. By incorporating information from temporal features, this dynamic kernel utilizes all frame features to enhance spatial patch representation. The two \texttt{FC} layers in \cref{eq:patch_cal} are also used to decrease the feature channels to accelerate the computations.
We then concatenate the [CLS] feature and patch feature to update the frame features $\tilde{\mathbf{v}} = [\tilde{\mathbf{v}}_{\rm [CLS]}, \tilde{\mathbf{v}}_{\rm patch}] \in \mathbb{R}^{N\times T\times d} $. 

The video adapter is inserted into the transformer block of the visual backbone. This process enriches each frame’s representation with video temporal features, thus enhancing the model’s effectiveness in aligning video and language features.

\noindent\textbf{Adaptive Training.}
As we have introduced uninitialized video adapter into the model, we need to task precautions to void damaging the existing pre-trained parameters of the CLIP model. To address this, we divide the training process into two steps: adaptive transferring and integrated tuning. Both stages involve the same pre-training tasks, as described in \cref{sec:pre-training-tasks} and data.
During the adaptive transferring stage, we freeze all parameters in CLIP except for those in the adapter part, which are trainable. This setting ensures that the model retains previously learned knowledge while also utilizing a carefully designed adapter module to learn new knowledge from a new dataset.
Once the first stage is complete, we move on to the second stage of pre-training the model. In the integrated tuning stage, we set all parameters in the model as trainable on the same dataset to further refine the model's performance. Detailed experiments are present in \cref{sec:va}.

\subsection{Feature Blending}
\label{sec:ensemble}

The encoder of the CLIP model prioritizes stationary content and spatial representations, while the video adapter is tailored to capturing dynamic contexts. Our method for utilizing both feature types involves two ensemble strategies for merging spatial and temporal features within the multi-modal encoder. The multi-modal encoder shares the same structure as in VLAB$_{\text{FA}}$: consisting of 12 blocks, each comprising sequential self-attention, cross attention, and feed forward network operations. We crafted following two cross-attention structures to execute the feature blending process.

\textbf{Stack.} This structure stacks two cross-attention layers for image and video features respectively. The stack composition strategy could be formulated as:
\begin{equation}
    \mathbf{x}_l^{ca} =\texttt{CA}^{v}(\texttt{CA}^{i}(\mathbf{x}_l^{sa};\mathbf{I});\mathbf{v})
    \label{eq:stack_composition}
\end{equation}
where, $\mathbf{x}_l^{ca}$ represents output feature of cross-attention structure in $l$-th block. $\mathbf{x}_l^{sa}$ represents output feature of self-attention in in $l$-th block. $\texttt{CA}^{v}$ and $\texttt{CA}^{i}$ represent cross-attention layers attend to video adapter features and image features. $\mathbf{I}$,$\mathbf{v}$ denote video features and image features respectively.

\textbf{Parallel.} We build two parallel cross-attention layers for image encoder and video adapter features respectively. Then, we add the two parallel cross-attention results as the final output. This strategy could be formulated as:
\begin{equation}
    \mathbf{x}_l^{ca} =\alpha \cdot \texttt{CA}^{v}(\mathbf{x}_l^{sa};\mathbf{v}) + \beta \cdot \texttt{CA}^{i}(\mathbf{x}_l^{sa};\mathbf{I})
    \label{eq:sum_composition}
\end{equation}
where $\alpha$, $\beta$ are learnable parameters, the learnable weights adaptively adjust the importance of two vision adapters.

By fusing the video features and image features through two cross-attention modules, the model's generative and contrastive learning capabilities can be effectively improved. In addition, we analyze the impact of the parameter sharing between the two cross-attention modules in~\cref{sec:feature_blending}.

\subsection{Pre-training Tasks}
\label{sec:pre-training-tasks}

Our goal is to learn a universal representation that can be utilized for diverse video and language tasks using self-supervised learning methods. To achieve this, we utilize video-text contrastive learning (VTC) to align visual and textual representations. Additionally, we incorporate masked language modeling (MLM) and unified language modeling (uni-LM) as pre-training tasks to enhance the model's language generation capabilities.
 
\noindent\textbf{Video-Text Contrastive Learning.}
Following~\cite{li2022blip,yu2022coca}, the contrastive loss function can be represented as: 
$\mathcal{L}_{vtc} = -\frac{1}{N} \sum_{i} log \frac{\text{exp}(\text{sim}(\mathbf{v}^{i},\mathbf{x}^{i})/\tau)}{\sum_j \text{exp}(\text{sim}(\mathbf{v}^{i},\mathbf{x}^{j})/\tau)}$
,where $\mathbf{v}^{i}$ and $\mathbf{x}^{i}$ are the paired image/video-text embeddings, and $\tau$ is a temperature parameter. sim(,) is the cosine similarity.

\noindent\textbf{Masked Language Modeling.} We follow~\cite{kenton2019bert} and adopt the Masked Language Modeling~(MLM) to enhance the model's language understanding capability, conditioned on visual representations. This loss function can be formulated as: 
$\mathcal{L}_{mlm} = - \frac{1}{N}\sum_{i} \sum_k log P(x_k^{i} | x^i; \mathbf{v}^{i})$
,where $x_k^{i}$ is the masked token. $\mathbf{v}^{i}$ is the visual representation.

\noindent\textbf{Unified Language Modeling.} 
To enhance the model's generative capability, we introduce Uni-LM~\cite{dong2019unified} as an additional training task alongside MLM. Both tasks share the same multi-modal encoder but differ in the attention pattern used in the Transformer. Uni-LM applies a causal attention mask, while MLM utilizes bi-directional attention. Uni-LM naturally adapts to sequential generation tasks based on preceding texts. Its formulation is: $\mathcal{L}_{uni-lm} = - \frac{1}{N}\sum_{i} \sum_k log P(x_k^{i} | x_{<k}^{i}; \mathbf{v}^{i})$
,where $x_k^{i}$ is the masked token, $x_{<k}^i$ is the sequence before $x_k^{i}$. $\mathbf{v}^{i}$ is the visual representation.

The final loss is calculated as the sum of the three aforementioned losses and utilized in the training of both VLAB$_{\text{FA}}$ and VLAB: 
\begin{equation}
\mathcal{L} = \mathcal{L}_{vtc} + \mathcal{L}_{mlm} + \mathcal{L}_{uni-lm}.
\end{equation}

\section{Experiments}

\subsection{Data and Evaluation Settings}

\noindent\textbf{Pre-training Datasets.} 
We use four public datasets in experiments. (a) \textit{CC4M} comprises a collection of datasets, including CC3M~\cite{sharma2018conceptual}, SBU~\cite{ordonez2011im2text}, Visual Genome~\cite{krishna2017visual} and COCO~\cite{chen2015microsoft}, amounting to 4M images or 5M image-text pairs. (b) \textit{CC12M}~\cite{changpinyo2021conceptual} dataset contains approximately 12M image-text pairs and is widely used for vision-language pre-training. (c)\textit{Webvid10M}~\cite{bain2021frozen} is a large and  comprehensive video dataset, consisting of 10.7M video-caption pairs. (d)\textit{Webvid2.5M}~\cite{bain2021frozen} is a subset of Webvid10M, used exclusively for conducting ablation experiments on the video adapter in our study.

\noindent\textbf{Downstream Tasks.}
\textit{Video Captioning.} Given a video, the model is require to generate a natual language description of the content. During fine-tuning stage, only the Uni-LM loss, as mentioned in \cref{sec:pre-training-tasks}, is applied to the multi-modal encoder. During inference, the captions are generated in an autoregressive manner.

\textit{Video Question Answering.} In VQA, the model receives a video and a question and predict the correct answer. We treat this task as a generation task, where during fine-tuing, we randomly mask the answer tokens and apply a uni-LM loss on the masked answer tokens and the end-of-sentence ([EOS]) token. During inference, the answer is generated based on the given video and question.

\textit{Text-Video Retrieval.} 
During fine-tuning, we discard the multimodal encoder and only utilize the contrastive loss for the visual and text encoders. Notably, our method offers superior practical efficiency compared to approaches that rely on video-text matching scores for retrieval, as it allows for offline extraction of video and text embeddings. We also use the dual softmax approach~\cite{cheng2021improving}.

\subsection{Implementation details}

To blend the adapter features, we first train the adapted model $\text{VLAB}_\text{FA}$ for learning representations from both temporal and spatial contexts. The adapter parameters are pre-trained on the video dataset Webvid10M. For the final model VLAB, two model configurations are presented: $\text{VLAB}_\text{L}$ and $\text{VLAB}_\text{G}$. 
$\text{VLAB}_\text{L}$ comprises 0.9B parameters and is constructed by incorporating the CLIP-large model and its adapted model, i.e. $\text{VLAB}_\text{FA}$. $\text{VLAB}_\text{G}$ replaces CLIP-large with EVA-CLIP-g~\cite{fang2022eva} while employing the same adapter model, resulting in a model with 1.6B parameters. The multi-modal encoder of both models is partly initialized from the BERT-base model~\cite{devlin2018bert}, which comprises 12 transformer blocks, whereas the rest of the cross-attention modules are randomly initialized. The VLAB model is optimized with the Adam optimizer~\cite{kingma2014adam}. The learning rate for the CLIP and Multi-Modal models are set to $5e^{-7}$ and $1e^{-4}$, respectively. We utilize stochastic depth regularization~\cite{huang2016deep} at a rate of 0.1 for video pre-training. We sparsely sample four frames from the video pre-training dataset and randomly crop both the image and video data at a resolution of $224\times224$. By default, we train both models for CC4M, CC12M, and Webvid10M on 64 V100 with a batch size of 2048 for 20/20/10 epochs. For other details, please refer to the Supplementary Material.

\subsection{Comparison to State-of-the-arts}
\label{sec:sota}

\begin{table*}[]
\small
\setlength{\tabcolsep}{12pt}
\centering
\caption{The results of video question answering on MSRVTT, MSVD and TGIF. VLAB sets new records across all datasets. Throughout the paper, we use \textbf{bold} and \underline{underline} to highlight the top two results.}
\begin{tabular}{l|c|ccc}
\toprule
    Method                    &\#Data(M)   & MSRVTT     & MSVD  &TGIF \\

    \midrule
    ALPRO~\cite{li2022align}            &5      &42.1   &45.9   &-       \\
    Clover~\cite{huang2022clover}       &5      &44.1   &52.4   &71.6   \\
    HiTeA~\cite{ye2022hitea}            &17     &45.9   &55.3   &73.2   \\
    OmniVL~\cite{wangomnivl}            &17     &44.1   &51.0   &-      \\
    mPLUG-2~\cite{xu2023mplug}          &17     &48.0   &58.1   &75.4   \\
    SINGULARITY~\cite{lei2022revealing} &17     &43.8   &-      &-       \\
    VINDLU~\cite{cheng2022vindlu}       &25    &44.6   &-      &-      \\
    LAVENDER~\cite{li2022lavender}      &30     &45.0   &56.6   &73.5   \\
    FrozenBiLM\cite{yang2022zero}       &30     &47.0   &54.8   &68.6   \\
    VideoCoCa~\cite{yan2022video}       &100   &46.3   &56.9   &-       \\
    All-in-one~\cite{wang2022all}       &283    &46.8   &48.3   &66.3   \\ 
    InternVideo~\cite{wang2022internvideo} &210  &47.1  &55.5   &72.2   \\ 
    GiT~\cite{wang2022git}              &800   &43.2   &56.8    &72.8      \\
    Flamingo~\cite{alayrac2022flamingo} &2300   &47.4   &-      &-      \\
    GiT2~\cite{wang2022git}             &12900  &45.6   &58.2   &74.9   \\ 
    \midrule
    $\text{VLAB}_{\text{L}}$ &26  &\underline{49.0} &\underline{59.2} &\underline{78.2}   \\
    $\text{VLAB}_{\text{G}}$ &26  &\textbf{49.6} &\textbf{61.0}       &\textbf{79.0}      \\
    \bottomrule
\end{tabular}

\label{tab:results_cap_qa}
\end{table*}

Following previous studies~\cite{wang2022git,li2022lavender}, we conduct thorough comparison with other state-of-the-art methods on a variety of highly competive benchmarks, including MSRVTT~\cite{xu2016msr}, MSVD~\cite{xu2017video}, Didemo~\cite{anne2017localizing}, and TGIF~\cite{jang2017tgif}.

\noindent\textbf{Open-ended Video QA.}
We evaluated our model on MSRVTT-QA, MSVD-QA, and TGIF-QA. The results are presented in \cref{tab:results_cap_qa}. $\text{VLAB}_{\text{G}}$ achieves accuracy of 49.6, 61.0, and 79.0 respectively. Despite using less pretrained data and having a smaller model size, $\text{VLAB}_{\text{L}}$ outperforms Flamingo and GiT by a significant margin. Remarkably, $\text{VLAB}_{\text{G}}$ establishes new records across all three competitive datasets. These results demonstrate the powerful capability of VLAB for complex multimodal understanding and reasoning.

\noindent\textbf{Video Captioning.}
As shown in \cref{tab:results_cap}, VLAB outperforms most methods and achieves state-of-the-art results on MSRVTT and MSVD datasets. In contrast to GiT2, which utilizes a significantly larger model (5.1B) and more extensive data (1.2B), our model achieves comparable performance. This highlights the superior efficiency of VLAB in learning video-language representations. It is worth noting that VLAB exhibits greater versatility and generality compared to GIT/GIT2, which only focus on generative tasks, while VLAB can also handle contrastive tasks like retrieval.

\noindent\textbf{Text Video Retrieval.}
Following previous works, we use the standard split of MSRVTT, MSVD, and Didemo datasets and report the retrieval results in terms of rank-1/5/10 and their summations. As shown in \cref{tab:video_retrieval}, our method achieves comparable results with less pre-training data compared to the previous SoTA method InternVideo, which also utilizes only contrastive loss. Note that InternVideo is specifically designed for video-language alignment and lacks the capability to perform generative tasks such as video captioning. Although the performance of VLAB is slightly inferior to methods (VindLU) that use matching loss/scores, we emphasize that ours is more suitable for practical use by enabling the extraction of video-text embeddings offline.

\begin{table*}[]
\small
\centering
\caption{The results of video captioning on MSRVTT and MSVD. VLAB is only inferior to GIT2 that employs much larger model (5.1B vs. 1.6B) and data (12800M vs. 26M). B@4: BLEU@4~\cite{papineni2002bleu}, M: METEOR~\cite{banerjee2005meteor}, R: ROUGE-L~\cite{lin2004rouge}, C: CIDEr~\cite{vedantam2015cider}.  All methods DO NOT perform any additional optimization like SCST.}
\begin{tabular}{l|c|cccc|cccc}
\toprule
\multirow{2}{*}{Method} & \multirow{2}{*}{\#Data(M)}  & \multicolumn{4}{c|}{MSRVTT}     & \multicolumn{4}{c}{MSVD} \\ \cmidrule(lr){3-10} 
                                                          &                    & B@4 & M & R & C & B@4 & M & R & C         \\ 
    \midrule
    SwinBERT~\cite{lin2022swinbert} & -        &41.9   &29.9   &62.1   &53.8   &58.2 &41.3 &77.5 &120.6 \\
    CLIP4Caption~\cite{tang2021clip4caption}  & &46.1 &30.7   &63.7   &57.7   &-  &-  &-  &-  \\
    HiTeA~\cite{ye2022hitea}       &17       &49.2   &30.7   &65.0   &65.1   &71.0   &45.3   &81.4   &146.9\\
    LAVENDER~\cite{li2022lavender}  &30     &-   &- &-   & 60.1  &-  &-  &- &150.7 \\  
    MV-GPT~\cite{seo2022end}        &69      &48.9   &\textbf{38.7}   &64.0   &60.0   &-  &-  &-  &-\\
    VideoCoca~\cite{yan2022video}     &100   &53.8 &- &68.0 &73.2 &- &- &- &- \\
    UniVL~\cite{luo2020univl}       &136     &42.2   &28.8   &61.2   &49.9   &-  &-  &-  &-\\
    GiT~\cite{wang2022git}          &800   & 53.8 & 32.9 & 67.7 &73.9   &79.5   &51.1   &87.3   &\underline{180.2} \\
    GiT2~\cite{wang2022git}         &12900  & \textbf{54.8} &33.1 & 68.2 &\textbf{75.9}   &\textbf{82.2}   &\textbf{52.3}   &\textbf{88.7}   &\textbf{185.4}\\ 
    
    \midrule
    $\text{VLAB}_{\text{L}}$       &26       &54.3   &32.7   &67.9   &72.5   &78.7   &50.3   &86.9   &174.1  \\
    $\text{VLAB}_{\text{G}}$       &26       &\underline{54.6}   &\underline{33.4}   &\textbf{68.3}   &\underline{74.9}   &\underline{79.3}   &\underline{51.2}   &\underline{87.9}   &179.8   \\
    \bottomrule
\end{tabular}
\label{tab:results_cap}
\end{table*}

\begin{table*}[]
    \small
    \setlength{\tabcolsep}{3.1pt}
    \centering
    \caption{The results of text-to-video retrieval on MSRVTT, MSVD and DIDEMO. Compared with InternVideo, VLAB achieves comparable performance while using much less data. VLAB is also efficient in practical usage. Refer to Sec.~\ref{sec:sota} for detailed explanations.}
    \begin{tabular}{l|c|cccc|cccc|cccc}
        \toprule
\multirow{2}{*}{Method} & \multirow{2}{*}{\#Data(M)}  & \multicolumn{4}{c|}{MSRVTT}     & \multicolumn{4}{c|}{MSVD} &\multicolumn{4}{c}{DIDEMO} \\ \cmidrule(lr){3-14} 
        &                    & R1 & R5 & R10 &SUM & R1 & R5 & R10 &SUM & R1 & R5 &R10  &SUM       \\ 
         \midrule
         \multicolumn{14}{l}{\textit{vison-text matching}}\\

         OmniVL~\cite{wangomnivl}           &18        &47.8 &74.2 &83.8 &205.8 &52.4 &79.5 &85.4 &217.3 &-\\
         LAVENDER~\cite{li2022lavender}     &30        &40.7 &66.9 &77.6 &185.2 &50.1 &79.6 &87.2 &216.9 & 53.4 &78.6 &85.3 &217.3 \\
         VIOLET~\cite{fu2021violet}         &186       &34.5 &63.0 &73.4 &170.9  &-     &-    &-    &- &32.6 &62.8 &74.7 &170.1\\
         VINDLU~\cite{cheng2022vindlu}                             &25       &48.8 &72.4 &82.2 &203.4     &-    &-    &-    &-   &59.8 &86.6 &91.5 &237.9 \\  
         mPLUG-2~\cite{xu2023mplug}                            &17        &53.1 &77.6 &84.7 &215.4     &-    &-    &-    &-  &56.4 &79.1 &85.2 &220.7 \\
         \midrule
         \multicolumn{14}{l}{\textit{vison-text contrastive}}\\
         CLIP4clip~\cite{luo2022clip4clip}  &-         &44.5 &71.4 &81.6 &197.5 &46.2 &76.1 &84.6 &206.9 &43.4 &70.2 &80.6 &194.2\\
         DCR~\cite{wang2022disentangled}    &-         &50.2 &76.5 &84.7 &211.4 &50.0 &81.5 &89.5 &221.0 &49.0 &76.5 &84.5 &210.0\\
         TS2Net~\cite{liu2022ts2}           &-         &49.4 &75.6 &85.3 &210.3 &41.8 &71.6 &82.0 &195.4 &-\\
         X-CLIP~\cite{ma2022x}              &-         &49.3 &75.8 &84.8 &209.9 &50.4 &80.6 &-    &-     &47.8 &79.3 &-    &- \\
         Clover~\cite{huang2022clover}                             &5         &40.5 &69.8 &79.4 &189.7     &-    &-    &-    &-     &50.1 &76.7 &85.6 &212.4 \\

         MV-GPT~\cite{seo2022end}              &69        &37.3 &65.5 &75.1 &177.9 &-    &-    &-    &- &-     &-    &-    &-\\
         TACo~\cite{yang2021taco}           &136       &28.4 &57.8 &71.2 &157.4 &-    &-    &-    &- &-     &-    &-    &-\\
         InternVideo~\cite{wang2022internvideo} &210   &\textbf{55.2} &\textbf{79.6}    &\underline{87.5}    &\textbf{222.3} &\textbf{58.4} &\textbf{84.5}    &\textbf{90.4}    &\textbf{233.3} &\textbf{57.9}  &\textbf{82.4}    &\textbf{88.9}    &\textbf{229.2}\\
         ALL-in-one~\cite{wang2022all}      &283       &37.9 &68.1 &77.1 &183.1 &-    &-    &-    &-  &32.7 &61.4 &73.5 &167.6\\

         \midrule
         $\text{VLAB}_{\text{L}}$ &26  &54.9 &78.6 &87.0 & 220.5 &55.4 &82.5 &89.2 &227.1 &55.1 &\underline{81.9} &87.6 &224.6  \\
         $\text{VLAB}_{\text{G}}$ &26 & \underline{55.1} & \underline{78.8} & \textbf{87.6} & \underline{221.5} &\underline{57.5} &\underline{83.6} &\underline{89.9} &\underline{231.1} &\underline{56.8} & 81.6 & \underline{88.7} &\underline{227.1}  \\
         \bottomrule
    \end{tabular}

    \label{tab:video_retrieval}
\end{table*}

\subsection{Method Analysis}

In this section, we conduct a series of ablation experiments to gain insights into the key components and validate the design choices adopted in VLAB. Next, we present the separate ablative results of feature adapting and feature blending, followed by a comprehensive comparison with a variant method to thoroughly verify the effectiveness of VLAB.

\subsubsection{Feature Adapting} 
\label{sec:va}

\begin{table}[t]
\centering
\begin{minipage}{.48\linewidth}
	\caption{Analysis of the data scalability of feature adapting. The results are obtained on MSR-VTT. W2M: Webvid2.5M, W10M: Webvid10M.}
\resizebox{\linewidth}{!}{
	\begin{tabular}{l|c|cccc}
	\toprule
		Method & Data            & Retrieval  & VQA  & Captioning \\
	\midrule
            Baseline  & W2M & 206.7 & 47.2  & 67.3 \\
            $\text{VLAB}_{\text{FA}}$ & W2M & 207.0(+0.3) & 47.4(+0.2) & 68.7(+1.4) \\ \midrule
            Baseline & W10M &207.1 &47.6 & 68.1 \\
		$\text{VLAB}_{\text{FA}}$ & W10M & 208.1(+1.0) & 48.0(+0.4) & 69.8(+1.7)  \\

	\bottomrule
	\end{tabular}
	}

	\label{tab:va_data}
\end{minipage}%
\hspace{1em}
\begin{minipage}{.48\linewidth}
	\caption{Analysis of the video adapter by varying the number of channels and layers. The results are obtained on MSR-VTT.}
\resizebox{\linewidth}{!}{
	\begin{tabular}{l|c|cccc}
	\toprule
		Channel & VA-layer            & Retrieval  & VQA  & Captioning \\
	\midrule
            $\text{VLAB}_{\text{FA}}-$64   & all & 207.9 & 47.8  & 69.4 \\
            $\text{VLAB}_{\text{FA}}-$64  & last 4 & 207.3 & 47.9 & 69.2 \\
		  $\text{VLAB}_{\text{FA}}-$64  & last 8 & 207.1 &  47.9 & 69.5 \\
            $\text{VLAB}_{\text{FA}}-$512 & last 4 & \textbf{208.1} & \textbf{48.0} & \textbf{69.8}  \\
	\bottomrule
	\end{tabular}
	}

	\label{tab:va_set}
\end{minipage}
\end{table}

\begin{table}[t]
\centering
\small
 \caption{Analysis of the training strategies for the video adapter. The Integrated Tuning denotes the model pre-trained with uninitialized adapter modules. The Adaptive Transferring is to tune the adapter modules first.}
\resizebox{1.0\linewidth}{!}{
\begin{tabular}{l|ccc|ccc|c}
	\toprule
		\multirow{2}{*}{Method}  & \multicolumn{3}{c|}{MSR-VTT} & \multicolumn{3}{c|}{MSVD} & \multicolumn{1}{c}{DiDeMo}
  
  \\ & Captioning  & VQA  & Retrieval &   Captioning  & VQA  & Retrieval & Retrieval \\
	\midrule
            Baseline  & 68.1 & 47.6 & 207.1 & 158.8 & 57.1 & 222.1 & 205.7 \\
             Integrated Tuning  & 69.5 & 47.8 & 207.2 & \textbf{168.1} & 58.1 & \textbf{222.6} & 203.8 \\
            Adaptive Transferring + Integrated Tuning  & \textbf{69.8} & \textbf{48.0} & \textbf{208.1} & 166.6 & \textbf{58.1} & 220.4 & \textbf{208.1}\\
	\bottomrule
	\end{tabular}
 }
 \label{tab:va_train}
\end{table}

\begin{table}[t!]
\centering
\small
 \caption{Analysis of feature blending from three perspectives: blending method, whether to share cross-attention weights, and whether to tune the adapted vision encoder.}
\resizebox{\linewidth}{!}{
\begin{tabular}{l|c|c|ccc|ccc}
	\toprule
		\multirow{2}{*}{Method} &\multirow{2}{*}{Unfreeze} &\multirow{2}{*}{Share}  & \multicolumn{3}{c|}{MSR-VTT} & \multicolumn{3}{c}{MSVD} 
  
  \\ &&& Captioning  & VQA  & Retrieval &   Captioning  & VQA  & Retrieval  \\
	\midrule
            Stack &- &-  & 71.1 & 48.7 & 210.0 &172.3  & 58.7    & 224.6 \\
            Stack  &-&\checkmark  &70.8 &48.8 &209.7 &171.1 & 58.8 & 224.1\\
            Stack &\checkmark &\checkmark &70.2 &48.7 &209.2 &170.4 &58.6 &224.7\\
            \hline
            Parallel &- &- &70.8 &48.7 & 210.0 &170.1 &58.6 & 224.2 \\
            Parallel &- &\checkmark & 70.9 & 48.8 &208.6 &171.2 &58.4 &224.2  \\
            Parallel &\checkmark &\checkmark &71.0 &48.7 & 209.2 & 170.7 &58.5 &224.7  \\
            
	\bottomrule
	\end{tabular}
 }

 \label{tab:ensemble}
\end{table}

\noindent\textbf{Effectivenss of Feature Adapting on the Pre-training Data Scale.}
We first study the scalability of video adapter on  video datasets of different sizes in \cref{tab:va_data}.  The baseline used in comparison shares the same configuration with VLAB$_{\text{FA}}$, excluding the use of the video adapter module. We observe that video adapter consistently improves the performance across all tasks and scales of data. Interestingly, the improvement is significant when trained on the large scale video dataset, which indicates that video adapter may be benefiting from learning more general features from the large-scale dataset. 

\noindent\textbf{Impact of Adapting Configuration.}
We then analyze the impact of different setups of the video adapter in \cref{tab:va_set}. We insert the video adapter in different layers and vary the number of channels. We observe that inserting the video adapter in the last four layers provides the best performance, and the performance drops as we use more adapter layers. Also, increasing the number of channels in the VA provides a slightly better performance. For the rest of experiments, the video adapter is inserted in the last four layers of the CLIP vision transformer.

\noindent\textbf{Strategy of Training the Adapter.}
The impact of our proposed training pipeline for adapting video features was assessed in \cref{tab:va_train}. We examined the efficacy of two training methods for the adapter: integrated tuning a pre-trained visual encoder with randomly initializing the video adapter, versus training the adapter while freezing the visual encoder and doing the integrated tuning later. Our findings suggest that an adaptively tuned video adapter consistently outperforms an integrally tuned version in all tasks, implying that training the video adapter first and then tuning both the visual encoder and adapter is preferred.

\subsubsection{Feature Blending}
\label{sec:feature_blending}
As shown in \cref{tab:ensemble}, we evaluate various configurations of feature blending on video-text tasks from three perspectives: blending method, weight sharing between cross attention, and tuning strategy. All models in the table are trained and evaluated under identical training settings. The training epochs are 10/10/5 for CC4M/CC12M/Webvid10M, respectively.

\noindent\textbf{Blending Method.} We evaluate two feature blending methods, namely ``stack'' and ``parallel'' as introduced in Sec.~\ref{sec:ensemble}. As it can be seen, both methods achieve comparable performance between in most tasks. For the subsequent experiments, we default to using the ``parallel'' method due to its efficiency potentials in both training and inference, as it enables parallel processing.

\begin{figure*}[!t]
\begin{center}
\includegraphics[width=\linewidth]{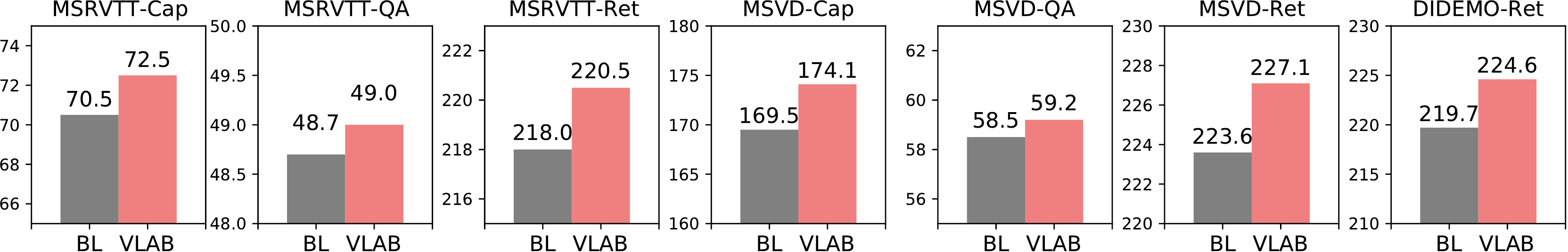}
\end{center}
   \caption{Comparison of VLAB with a well-established baseline method on three downstream tasks. ``Cap/QA/Ret'' denote the video captioning/qa/retrieval tasks respectively.}
\label{fig:improvement}
\end{figure*}

\noindent\textbf{Sharing Between Cross Attention.}
Recall that we utilize a pair of cross-attention layers in the multi-modal encoder to merge temporal and spatial visual features. To assess the impact of weight sharing in cross-attention, we compare its effect on the final performance of VLAB in Tab.~\ref{tab:ensemble}. We observe that sharing the cross-attention parameters between the two types of visual features proves to be sufficiently effective in video-text tasks. Moreover, considering it is more memory-friendly, we thus use the sharing way in experiments.

\noindent\textbf{Tuning Strategy.} 
We further examine the impact of freezing or unfreezing the adapted vision encoder during feature blending training. Due to GPU memory limitations, we can only unfreeze the last 4 layers. As shown in Tab.~\ref{tab:ensemble}, we observe that further training of the unfrozen layers does not result in significant performance improvement. Based on these findings, we choose to freeze the adapted encoder to enhance training efficiency and reduce memory usage.

\subsubsection{Comparison to a variant method}

To further demonstrate the effectiveness of our proposed methods in VLAB, we establish a strong baseline for comparison. Specifically, we remove the adapted vision encoder depicted in Fig.~\ref{fig:frame_work}(c) while keeping the remaining components and training loss the same as in VLAB. The comparison results between these two approaches are illustrated in Fig.~\ref{fig:improvement}. As one can observe, VLAB consistently outperforms this variant method across all tasks. These results clearly evidence the advantages of the proposed method in enhancing video-language representation learning.

\section{Conclusion}

In this paper, we present VLAB, a novel video language pre-training method that leverages learned representations from image-text models to enhance video-text understanding. Benefiting from the proposed feature adapting and blending approaches, VLAB generates high-quality video language representations and handles both generative and contrastive video language tasks within a unified framework. Through extensive experiments, VLAB achieves competitive results across various benchmarks. In the future, we plan to scale up our method by using larger datasets and models.

\bibliographystyle{IEEEtran}
\small
\bibliography{reference}

\end{document}